\newif\ifanonymized\anonymizedfalse  
\newif\ifroughdraft\roughdraftfalse  
\newif\ifdraft\draftfalse  
\newif\ifdraft\drafttrue  

\RequirePackage[l2tabu,orthodox]{nag}

\RequirePackage{iftex}
\RequirePDFTeX

\documentclass[conference\ifroughdraft,draft\fi]{IEEEtran}

\usepackage[utf8]{inputenc}
\usepackage[T1]{fontenc}
\usepackage[USenglish]{babel}
\usepackage{lmodern}   
\usepackage{newtxtext} 
\usepackage{amsmath}   
\usepackage{amsfonts}
\usepackage{amssymb}
\usepackage{newtxmath} 
\usepackage[final]{microtype} 

\usepackage[noend,ruled,linesnumbered]{algorithm2e}
\usepackage{booktabs}
\usepackage{cite}
\usepackage{comment}
\usepackage{enumitem}
\usepackage{multirow}
\usepackage[final]{graphicx}
\usepackage{xcolor}
\usepackage{xspace}

\usepackage[hyphens]{url}
\usepackage[
bookmarksnumbered,
colorlinks=false,
final,
hidelinks,
pdfa,
pdflang=en,
pdftex,
unicode
]{hyperref}
\usepackage{cleveref}
\usepackage{bookmark}

\makeatletter

\def\@IEEEconsolenoticeconference{}

\makeatother

\AtBeginDocument{%
	\sfcode\csname\encodingdefault\string\textquotedblright\endcsname=0%
	\sfcode\csname\encodingdefault\string\textquoteright\endcsname=0%
}

\DeclareMicrotypeSet[protrusion]{not-tt}{
	family = {rm*, sf*}
}
\UseMicrotypeSet[protrusion]{not-tt}

\newlength{\interwordspace}
\newcommand{\etal}{et~al.\xspace} 
\newcommand{\etc}{etc.\xspace} 
\newcommand{\ie}{i.e.\xspace} 
\newcommand{\eg}{e.g.\xspace} 
\newcommand{\nop}[1]{}

\ifdraft
\newcommand{\kibitz}[3]
{\textcolor[HTML]{#1}{[\textbf{#2}\ifx&#3&\else: \textit{#3}\fi]}}
\else
\newcommand{\kibitz}[3]{}
\fi

\title{Self-Supervised Euphemism Detection and Identification for Content Moderation}
\IEEEoverridecommandlockouts
\author{%
	\ifanonymized
	\\ 
	\IEEEauthorblockN{Anonymous Authors}
	\IEEEauthorblockA{
		\textit{Anonymous Institutions}\\ \\ 
	}
	\else
	
	Wanzheng Zhu$^{*}$, Hongyu Gong$^{\dagger*}$\thanks{$^{\dagger}$ The work was done while Hongyu Gong was at UIUC.}, Rohan Bansal$^{\ddag}$, 
	Zachary Weinberg$^{\S\ddag}$,\\ 
	Nicolas Christin$^{\ddag}$, Giulia Fanti$^{\ddag}$, and Suma Bhat$^{*}$\\
	\normalsize
	$^{*}$University of Illinois at Urbana-Champaign, 
	$^{\dagger}$Facebook,\\
	$^{\ddag}$Carnegie Mellon University, 
	$^{\S}$University of Massachusetts, Amherst\\
	$^{*}$\{wz6, spbhat2\}@illinois.edu, $^{\dagger}$hygong@fb.com \\
	$^{\ddag}$\{rohanb, nicolasc, gfanti\}@andrew.cmu.edu, $^{\S}$zackw@cs.umass.edu
	\fi}
\date{November 2020}

\begin{document}
	\maketitle
	\begin{abstract}
		Fringe groups and organizations
		have a long history of using \emph{euphemisms}---%
		ordinary-sounding words with a secret meaning---%
		to conceal what they are discussing.
		Nowadays, one common use of euphemisms
		is to evade content moderation policies
		enforced by social media platforms.
		Existing tools for enforcing policy automatically
		rely on keyword searches for words on a ``ban list'', 
		but these are notoriously imprecise:
		even when limited to swearwords,
		they can still cause embarrassing false positives~\cite{blackie1996:scunthorpe}.
		When a commonly used ordinary word acquires a euphemistic meaning,
		adding it to a keyword-based ban list is hopeless:
		consider ``pot'' (storage container or marijuana?)
		or ``heater'' (household appliance or firearm?)
		The current generation of social media companies
		instead hire staff to check posts manually,
		but this is expensive, inhumane, and not much more effective.
		It is usually apparent to a human moderator
		that a word is being used euphemistically,
		but they may not know what the secret meaning is,
		and therefore whether the message violates policy.
		Also, when a euphemism is banned,
		the group that used it need only invent another one,
		leaving moderators one step behind.
		
		This paper will demonstrate unsupervised algorithms that,
		by analyzing words in their sentence-level context,
		can both detect words being used euphemistically,
		and identify the secret meaning of each word.
		Compared to the existing state of the art,
		which uses context-free word embeddings,
		our algorithm for detecting euphemisms
		achieves 30--400\% higher detection accuracies
		of \emph{unlabeled} euphemisms in a text corpus.
		Our algorithm for revealing euphemistic meanings of words
		is the first of its kind,
		as far as we are aware.
		In the arms race between content moderators and policy evaders,
		our algorithms may help shift the balance in the direction of the moderators. 
	\end{abstract}
	
	\begin{IEEEkeywords}
		Euphemism detection, Euphemism identification, Self-supervised learning, Masked Language Model (MLM), Coarse-to-fine-grained classification
	\end{IEEEkeywords}
	
	\section{Introduction}
	\label{sec:intro}
	
	In recent years, large social media companies have been hiring content moderators to prevent conversations on their platforms that they deem to be inappropriate.
	Even though content moderation---the process of deciding what stays online and what gets taken down---often relies on organization-wide, centralized policies, 
	the people who do this job often feel marginalized~\cite{barrett2020moderates}. 
	In 2019, The Verge reported on the emotional toll this work exacts,
	leading in some cases
	to post-traumatic stress disorder~\cite{TheVerge:Mods, TheVerge:Mods2}.

	
	Automation is an obvious way to assist content moderators.
	Ideally, they would be able to make a decision once
	and have it applied consistently to all similar content.
	One standard form of automated moderation is ``ban-lists'' of forbidden words.
	These are easy to implement, and define a clear-cut policy.
	However, they are also easy to evade:
	as soon as terms are added to a ban-list,
	the offenders will notice
	and adapt by inventing euphemisms to evade the filters~\cite{Ofcom:AI2019}.
	Euphemisms are frequently words with other, innocuous meanings
	so they cannot be filtered unconditionally;
	they must be interpreted in context.
	To illustrate the problem,
	Table~\ref{table:example2} gives many examples of euphemisms
	for a few terms that are frequently forbidden.
	Almost all of the euphemisms have innocuous meanings.
	Table~\ref{table:example1} shows how a few of the euphemisms
	would be used in context, demonstrating that
	a human reader can often tell that a euphemistic meaning is intended
	even if they do not know exactly what the meaning is.
	
	We present techniques for automated assistance
	with two tasks related to ban-list maintenance.
	Our algorithm for \textbf{euphemism detection}
	takes as input a set of \textit{target keywords} referring to forbidden topics
	and produces a set of \textit{candidate euphemisms}
	that may signify the same concept as one of the target keywords,
	without identifying which one.
	\textbf{Euphemism identification} takes a single euphemism as input
	and identifies its meaning.
	We envision these algorithms being used in a pipeline
	where moderators apply both in succession
	to detect new euphemisms and understand their meaning.
	For instance, if the target keywords are formal drug names 
	(\eg, marijuana, heroin, cocaine),
	euphemism detection might find common slang names for these drugs
	(\eg, pot, coke, blow, dope)
	and euphemism identification could then associate each euphemism
	with the corresponding formal name
	(\eg, $\text{pot} \longrightarrow \text{marijuana}$,
	$\text{coke, blow} \longrightarrow \text{cocaine}$,
	$\text{dope} \longrightarrow \text{heroin}$).
	
	In addition to their practical use in content moderation,
	our algorithms advance the state of the art in Natural Language Processing (NLP)
	by demonstrating the feasibility of self-supervised learning
	to process large corpora of unstructured, non-canonical text
	(\eg, underground forum posts),
	a challenging task of independent interest to the NLP community
	(\eg, \cite{durrett2017identifying,portnoff2017tools,felbo2017using}).
	Our algorithms require no manual annotation of text,
	and do not just rely on a ``black box''  pre-trained and
	fine-tuned model.
	
	\begin{table*}[t!]
		\centering
		\small
		\caption{Examples of the variety of euphemisms associated with target keywords in commonly forbidden categories.}
		\begin{tabular}{lll}
			\toprule
			\multicolumn{1}{c}{\textbf{Category}} & \multicolumn{1}{c}{\textbf{Target Keyword}} & \multicolumn{1}{c}{\textbf{Euphemisms}} \\
			\midrule
			& Marijuana    & blue jeans, blueberry, grass, gold, green, kush, popcorn, pot, root, shrimp, smoke, sweet lucy, weed \\
			\textbf{Drugs}   & Methamphetamine & clear, dunk, gifts, girls, glass, ice, nails, one pot, shaved ice, shiny girl, yellow cake \\
			& Heroin       & avocado, bad seed, ballot, beast, big H,  cheese, chip, downtown, hard candy, mexican horse, pants \\
			\addlinespace
			\textbf{Weapons} & Gun          & bap, boom stick, burner, chopper, cuete, gat, gatt, hardware, heater, mac, nine, piece, roscoe, strap \\
			& Bullet       & ammo, cap, cop killer, lead, rounds \\
			\addlinespace
			\textbf{Sex}     & Breasts      & bazooms, boobs, lungs, na-nas, puppies, tits, yabo \\
			& Prostitution & call girl, girlfriend experience, hooker, poon, whore, working girl \\
			\bottomrule
		\end{tabular}
		\label{table:example2}
	\end{table*}
	
	\begin{table*}[ht!]
		\centering
		\small
		\caption{Example usage for a few of the euphemisms in Table~\ref{table:example2}.}
		\begin{tabular}{l@{\hspace{\interwordspace}}l@{\qquad}l}
			\toprule
			& \multicolumn{1}{c}{\textbf{Example Sentences} (euphemism in boldface)}
			& \multicolumn{1}{c}{\textbf{Euphemism means}} \\
			\midrule
			1. & I had to shut up: the dealers had \textbf{gats}, my boys didn't.
			& \textit{machine pistol} \\\addlinespace[3pt]
			2. & For all vendors of \textbf{ice},
			it seems pretty obvious that it is not as pure as they market it.
			& \textit{methamphetamine} \\\addlinespace[3pt]
			3. & I feel really good and warm behind the eyes.
			It's not something I've felt before on \textbf{pot} alone to this degree.
			& \textit{marijuana} \\\addlinespace[3pt]
			4. & You can get an ounce of this \textbf{blueberry kush}
			for like \$300 and it's insane.
			& \textit{variety of marijuana} \\\addlinespace[3pt]
			5. & I'm looking for the \textbf{girlfriend experience},
			without having to deal with an actual girlfriend.
			& \textit{form of prostitution} \\
			\bottomrule
		\end{tabular}
		\label{table:example1}
	\end{table*}
	
	\begin{table}[ht]
		\centering
		\small
		\caption{Example informative and uninformative contexts.
			\upshape
			The word ``heroin'' has been masked out of each sentence below. In cases 1--3 it is clear that the masked word must be the name of an addictive drug, while in cases 4--6 there are more possibilities.}
		\newcommand{\maskedword}{\rule{3em}{1.5ex}}
		\begin{tabular}{ll@{\hspace{\interwordspace}}p{0.64\columnwidth}}
			\toprule
			\multicolumn{1}{c}{\textbf{Context}} &  \multicolumn{2}{c}{\textbf{Example Sentences}} \\
			\midrule
			\textbf{Informative}
			& 1. & This 22 year old former \maskedword{} addict who I did drugs with was caught this night. \\
			& 2. & I have xanax real roxi opana cole and \maskedword{} for sale. \\
			& 3. & Six \maskedword{} overdoses in seven hours in wooster two on life support. \\\addlinespace
			\textbf{Uninformative}
			& 4. & Why is it so hard to find \maskedword{}? \\
			& 5. & The quality of this \maskedword{} is amazing and for the price its unbelievable. \\
			& 6. & Could we in the future see \maskedword{} shampoo? \\
			\bottomrule
		\end{tabular}
		\label{table:example3}
	\end{table}
	
	\subsection{Euphemism Detection}
	The main challenge of automated euphemism detection
	is distinguishing the euphemistic meaning of a term
	from its innocuous ``cover'' meaning~\cite{yuan2018reading}.
	For example, in sentence 2 of Table~\ref{table:example1},
	``ice'' \emph{could} refer to frozen water. 
	To human readers, this is unlikely in context,
	because the purity of frozen water
	is usually not a concern for purchasers.
	Previous attempts to automate this task~\cite{takuro2020codewords,magu2018determining,yuan2018reading,zhao2016chinese}
	relied on \emph{static word embeddings} (\eg, word2vec~\cite{mikolov2013distributed,mikolov2013efficient}),
	which do not attempt to distinguish different senses of the same word.
	They can identify slang terms with only one meaning
	(\eg, ``ammo'' for bullets),
	but perform poorly on euphemisms.
	Continuing the ``ice'' example,
	sentences using it in its frozen-water sense
	crowd out the sentences using it as a euphemism
	and prevent the discovery of the euphemistic meaning.
	
	A newer class of \emph{context-aware} embeddings (\eg BERT~\cite{devlin2019bert})
	learns a different word representation
	for every context in which the word appears,
	so they do not conflate different senses of the same word.
	However, since there are now several vectors associated with each word,
	the similarity of two words is no longer well-defined.
	This means context-aware embeddings cannot be substituted
	for the static embeddings used in earlier euphemism detection papers,
	which relied on word similarity comparisons.
	Also, not all contexts are equal.
	For any given term,
	some sentences that use it
	will encode more information about its meaning
	than others do.
	Table~\ref{table:example3} illustrates the problem:
	it is easier to deduce what the masked term probably was
	in sentences 1--3
	than sentences 4--6.
	This can be addressed by manually labeling sentences
	as informative or uninformative,
	but our goal is to develop an algorithm that needs no manual labels.
	
	In this paper, we design an end-to-end pipeline for detecting euphemisms by making \textit{explicit} use of context. 
	This is particularly important to help content moderation of text in forums. 
	We formulate the problem as an unsupervised fill-in-the-mask problem \cite{devlin2019bert,donahue2020enabling} and solve it by combining a masked language model (\eg, used in BERT \cite{devlin2019bert}) with a novel self-supervised algorithm to filter out uninformative contexts.
	The salience of our approach, which sets itself apart from other work on euphemism detection, lies in its non-reliance on linguistic resources (\eg, a sentiment lexicon) \cite{felt2020recognizing}, search-engine results, or a seed set   of euphemisms. 
	As such it is particularly relevant to our application case---online
	platforms with free-flowing discourse that may adopt their own 
	vernacular over time. 
	Evaluating on a variety of representative datasets of online posts
	we found that our approach yields  top-$k$ detection accuracies that are 30--400\% higher than state-of-the-art baseline approaches on all of the datasets, with top-20 accuracies as high as 40--45\%, which is high for this problem. 
	A qualitative analysis reveals that our approach also discovers correct euphemisms that {\em were not on our ground truth lists}, \ie, it can 
	detect previously unknown euphemisms. 
	Again, this is highly valuable in the context of 
	Internet communities, where memes 
	and slang lead to rapidly evolving vocabulary.
	
	\subsection{Euphemism Identification}
	\label{sec:intro_iden}
	\begin{table*}[t!]
		\centering
		\small
		\caption{Example uses of words in both euphemistic and non-euphemistic senses. All sentences are from Reddit.}
		\begin{tabular}{lll}
			\toprule
			\textbf{Word} & \textbf{Meaning} & \multicolumn{1}{c}{\bfseries Sentences} \\
			\midrule
			\multirow{8}{*}{Coke} & \multirow{4}{*}{Cocaine}
			& We had already paid \$70 for some shitty weed from a taxi driver
			but we were interested in some coke \\
			&   & \quad and the cubans. \\
			&	& Why are coke dealers the most nuttiest? \\
			&	& OK so we have one gram high quality coke
			between 2 people who have never done more than a bump. \\
			\addlinespace
			& \multirow{3}{*}{Coca-Cola}
			& I love having coke with ice. \\
			&	& When I buy coke at the beverage shop in UK,
			I pay neither a transaction fee nor an exchange fee. \\
			&	& Never have tried mixing coke with sprite or 7up. \\ 
			\midrule
			\multirow{8}{*}{Pot} & \multirow{4}{*}{Marijuana}
			& My cousin did the same and when the legalized pot in dc
			they really started cracking down \\
			&   & \quad in virginia and maryland. \\
			&	& As far as we know he was still smoking pot but that was it. \\
			&	& Age 17, every time I smoked pot, I felt out of place. \\
			\addlinespace
			& \multirow{3}{*}{Container}
			& No one would resist a pot of soup. \\
			&	& There's plenty of cupboard space in the kitchen
			for all your pots and pans. \\
			&	& Most lilies grow well in pots. \\
			\bottomrule
		\end{tabular}
		\label{table:example4}
	\end{table*}
	
	Once the usage of euphemisms has been detected, it is important to \emph{identify} what each euphemism refers to. Unlike the  task of deciding whether a given word refers to \textit{any} target keyword (euphemism detection), the task of euphemism identification maps a given euphemism to a \textit{specific} target keyword. This involves not only using the nuance of contextual information but also aggregating this information from related instances across the collection to make the inference.     
	Again, referring to the 2nd and 3rd examples in Table \ref{table:example1}, we want to identify that {\em ice} refers to {\em methamphetamine} and {\em pot}  to {\em marijuana}. 
	To the best of our knowledge, no prior work has explicitly captured the meaning of a euphemism except for a few peripheral works (\eg, \cite{yuan2018reading}) that identify the broad category of a euphemism (\eg, sedative, narcotic, or stimulant for a drug euphemism). 
	
	Euphemism identification poses four main challenges: 
	
	1) The distinction in meaning between the target keywords  (\eg, cocaine and marijuana) is often subtle and difficult  to learn from  raw text corpora alone. 
	2) A given euphemism can  be used in a euphemistic or  non-euphemistic sense, adding  the extra layer of linguistic nuance (Table \ref{table:example4}). 
	3) No curated datasets that are publicly available are adequate to exhaustively learn a growing list of mappings between euphemisms and their target keywords.
	4) It is unclear what linguistic and ontological resources one would need to automate this task. 
	
	In this paper, we propose the first approach to identify the precise meaning of a euphemism (\eg, mapping {\em pot}  to {\em marijuana} and {\em Adam} to {\em ecstasy}). 
	We systematically address  the  challenges identified above via a self-supervised learning scheme, a classification formulation, and a coarse-to-fine-grained framework. 
	The key novelty lies in how we formulate the problem and solve it without additional resources or supervision. 
	Going beyond demonstrating the feasibility of the task on a variety of datasets, we observe improvements in top-$k$ accuracy between 25--80\%  compared to constructed baseline approaches.

	
	\section{Related Work}
	\label{sec:related_work}
	Natural language processing (NLP) 
	has been used effectively in various security and privacy problems, including clustering illicit online pharmacies \cite{leontiadis2011measuring,mccoy2012pharmaleaks}, identifying sensitive user inputs \cite{huang2015supor,nan2015uipicker}, and detecting spam \cite{thomas2013trafficking,sedhai2017semi,wu2018twitter,wu2017twitter}. 
	However, although euphemisms have been widely studied in linguistics and related disciplines  \cite{keith1991euphemism,pfaff1997metaphor,hugh2002rawson,allan2009connotations,rababah2014translatability,spears1981slang,chilton1987metaphor,ahl2006motivation,fernandez2006language}, they have  received relatively little attention from the NLP \cite{felt2020recognizing}, or security and privacy communities. 
	Next, we review relevant prior work, including: 1) euphemism detection, 
	2) euphemism identification, and 3) self-supervised learning. 

	\subsection{Euphemism Detection}
	\label{sec:related_det}
	
	\begin{table*}[t!]
		\centering
		\small
		\caption{Related work on euphemism detection.}
		\begin{tabular}{p{0.10\textwidth}p{0.12\textwidth}p{0.22\textwidth}p{0.22\textwidth}p{0.22\textwidth}}
			\toprule
			\centering{\textbf{System}} & \centering{\textbf{Learning Type}} & \centering{\textbf{Categories (Platform)}} & \centering{\textbf{Required Input}} & \multicolumn{1}{c}{\textbf{Approach Keywords}} \\
			\midrule
			\textbf{Durrett \etal (2017) \cite{durrett2017identifying}} & Supervised \& semi-supervised  & Cybercriminal wares (Darkode), cybersecurity (Hack Forums), search engine optimization techniques (Blackhat), data stealing tools and services (Nulled) & A fully labelled dataset with annotated euphemisms & Support Vector Machine (SVM), Conditional Random Field (CRF) \\
			\\ 
			\textbf{Pei \etal (2019) \cite{pei2019slang}} & Supervised  & General topics (Online Slang Dictionary)  & Slang-less corpus (Penn Treebank) as the negative examples, Slang-specific corpus (Online Slang Dictionary) as the positive examples & Linguistic features, bidirectional LSTM \cite{huang2015bidirectional} , Conditional Random Field (CRF) \cite{lafferty2001conditional}, multilayer perceptron (MLP) \cite{rauber2011kernel} \\ 
			\\
			\textbf{Zhao \etal (2016) \cite{zhao2016chinese}} & Unsupervised  & Cybersecurity (QQ) & Target keywords, online search service & Unsupervised learning, word embedding (\ie, word2vec), Latent Dirichlet Allocation (LDA) \\
			\\
			\textbf{Yang \etal (2017) \cite{yang2017learn}} & Unsupervised  & Sex, gambling, dangerous goods, surrogacy, drug, faked sites (Baidu) & Target keywords, online search service & Web analysis, keywords expansion, candidate filtering\\
			\\
			\textbf{Hada \etal (2020) \cite{takuro2020codewords}} & Unsupervised & Drug trafficking and enjo kosai (Twitter) & A clean background corpus, a bad corpus related to illegal transactions, a set of euphemism seeds & Word embedding (word2vec), cosine similarity \\
			\\
			\textbf{Felt \etal (2020) \cite{felt2020recognizing}} & Unsupervised & Firing, lying and stealing (The English Gigaword corpus) & Category name, a lexicon dictionary (\ie, Gigaword) & Sentiment analysis, bootstrapping, semantic lexicon induction \\
			\\
			\textbf{Taylor \etal (2017) \cite{taylor2017surfacing}} & Unsupervised &  Hate speech (Twitter) &  The text corpus, category name & Word embedding (fasttext \cite{bojanowski2017enriching} and dependency2vec \cite{levy2014dependency}), community detection, bootstrapping \\
			\\
			\textbf{Magu \etal (2018) \cite{magu2018determining}} & Unsupervised &  Hate speech (Twitter) & The text corpus, a euphemism seed & Word embedding (word2vec), network analysis, centrality measures \\
			\\
			\textbf{Yuan \etal (2018) \cite{yuan2018reading}} & Unsupervised & Sale and trade of hacking services and tools (Darkode), blackhat hacking (Hack Forums), data stealing tool and service (Nulled), illegal drug (Silk Road) & A background corpus (\eg, Wikipedia), A dark corpus (\eg, Silk Road \cite{christin2013traveling}), A mixed corpus (\eg, Reddit) & Word embedding, semantic comparison across corpora \\
			\midrule
			\textbf{Our algorithm} & Unsupervised & Drug (Reddit), weapon (Gab, SlangPedia, \cite{durrett2017identifying,portnoff2017tools}), sexuality (Gab) & The text corpus, target keywords & Contextual information, masked language model, BERT \\
			\bottomrule
		\end{tabular}
		\label{table:related_dec}
	\end{table*}

	Euphemism detection is broadly related to the tasks of set expansion \cite{shen2017setexpan,zhu2019fuse,zhang2020empower,huang2020guiding,shen2020synsetexpan,rong2016egoset} and lexicon construction and induction \cite{hamilton2016inducing,yang2020co,huang2020corel,mao2020octet,shang2020nettaxo,shen2020taxoexpan,zhang2018taxogen}. 
	Set expansion aims to expand a small set of seed entities into a complete set of relevant entities, and its goal is to find other target keywords from the same category. 
	Lexicon construction and induction focus on extracting relations and building the lexicon-based knowledge graph in a structured manner. 
	Their goals are different from ours, which is to find euphemisms of target keywords.

	The specific task of euphemism detection has been studied in the NLP literature under a number of frameworks, including supervised, semi-supervised, and unsupervised learning, summarized in Table \ref{table:related_dec}.
	For example, Yang et al.\ \cite{yang2017learn} build a  Keyword Detection and Expansion System (KDES) and apply it to the search results of Baidu, China's top search engine. 
	KDES aims to infer whether a search keyword should be blocked by inspecting the associated search results.
	This approach requires general domain information with distant-supervision (\ie, the Baidu search engine), and is therefore not suitable for our unsupervised setting. 
	Even if assuming search engine access, 
	euphemisms for sensitive keywords are often short and innocent-looking (\eg, blueberries), 
	which may result in mainly legitimate search results. 
	Another set of relevant 
	articles \cite{durrett2017identifying,portnoff2017tools} generate high-level information to analyze underground forums via an automated, 
	top-down approach that blends information extraction and named-entity recognition. 
	They present a data annotation method and utilize the labeled data to train a supervised learning-based
	classifier. 
	Yet, the results depend heavily on the quality of annotation, and as shown by several researchers \cite{durrett2017identifying,yuan2018reading}, the model does not perform as well in cross-domain datasets, 
	where it is outperformed by standard  semi-supervised learning techniques. 
	
	\begin{figure*}[t]
		\centering
		\includegraphics[width=1.00\linewidth]{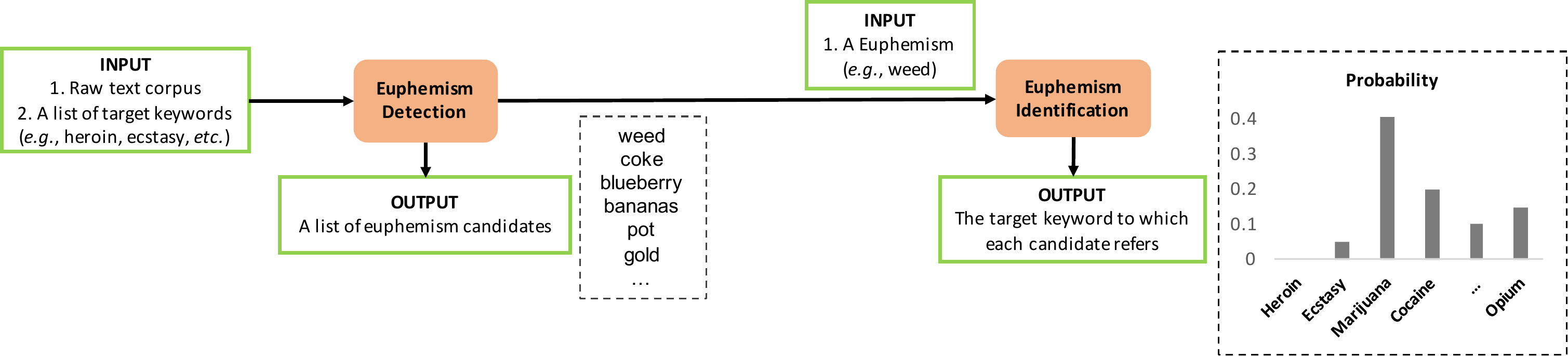}
		\caption{Euphemism detection and identification pipeline.}
		\label{fig:model_overview}
	\end{figure*}
	
	Our work is most closely related to four state-of-the-art approaches \cite{yuan2018reading,felt2020recognizing,taylor2017surfacing,magu2018determining}. 
	CantReader \cite{yuan2018reading} aims to automatically identify ``dark jargon'' from cybercrime marketplaces. 
	CantReader employs a neural-network based embedding technique to analyze the semantics of words, and detects euphemism candidates whose contexts in the background corpus (\eg, Wikipedia) are significantly different from those in the target corpus. 
	Therefore, it takes as input a ``dark'' corpus (\eg, Silk Road anonymous online marketplace \cite{christin2013traveling} forum), a mixed corpus (\eg, Reddit), and a benign corpus (\eg, English Wikipedia). 
	Different from CantReader, 
	we assume only access to a single target corpus -- although we do rely on context-aware embeddings that could be pre-trained from a reference corpus like Wikipedia, and then fine-tuned to the target corpus.
	More importantly, we find that our approach outperforms CantReader, presumably because we explicitly use context.
	
	Another relevant baseline \cite{felt2020recognizing} detects euphemisms instead by using sentiment analysis. 
	It identifies a set of  euphemism candidates using a bootstrapping algorithm for semantic lexicon induction. 
	Though the methodology seems reasonable and intuitive at first, it requires additional manual filtering process to refine the candidates and thus, fails to meet the requirement of automatic, large-scale detection 
	that online content moderators desire. 
	In yet another approach, Magu et al.\ \cite{magu2018determining} and Taylor et al.\ \cite{taylor2017surfacing} propose two algorithms that leverage word embeddings and community detection algorithms. 
	Magu et al.\ \cite{magu2018determining} generates a cluster of euphemisms by the ranking metric of eigenvector centralities \cite{bonacich1972factoring,bonacich1972technique}.
	Due to the intrinsic nature of the algorithm, this approach 
	requires a starting euphemism seed to find others. 
	Taylor et al.\ \cite{taylor2017surfacing} creates neural embedding models that capture the word similarities, uses graph expansion and the PageRank scores \cite{page1999pagerank} to bootstrap initial seed words, and finally enriches the bootstrapped words to learn out-of-dictionary terms that behave like euphemisms.
	However, the approaches of Magu et al.\ \cite{magu2018determining} and Taylor et al.\ \cite{taylor2017surfacing} were tested on one single dataset.  
	Unfortunately, 
	we do not find their performance to be as strong on the multiple datasets we evaluate. 
	

	\subsection{Euphemism Identification}
	\label{sec:related_iden}
	To the best of our knowledge, no work has explicitly attempted to infer euphemism meaning.
	Yuan et al.\ \cite{yuan2018reading} tackles a related problem by identifying the hypernym of euphemisms (\eg, whether it refers to a drug or a person).
	In a more general sense, the task of euphemism identification is also related to sense discovery of unknown words \cite{ishiwatari2019learning,ni2017learning} and word sense disambiguation \cite{taghipour2015semi,raganato2017neural,raganato2017word,iacobacci2016embeddings}. 
	While sense discovery aims to understand the meaning of an unknown word by generating a definition sentence,  
	word sense disambiguation focuses on identifying which sense of a word is used in a sentence, given a set of candidate senses and relies heavily on a sense-tagged reference corpus, created by linguists and lexicographers.
	However, neither of these are able to capture nuanced differences between a group of semantically-similar target keywords in the same category. 
	
	\subsection{Self-supervised Learning}
	The technical innovations in our work rely heavily on \emph{self-supervision}, a form of unsupervised learning where the data itself provides the supervision \cite{weng2019selfsup}. 
	Self-supervision was designed to make use of vast amounts of unlabelled data (\eg, free text, images) 
	by constructing a supervised learning task from the data itself to predict some attribute of the data.
	For example, to train a text prediction model, one can take a corpus of text, mask part of the sentence, and train the model to predict the masked part;
	this workflow creates a supervised learning task from unlabelled data.
	Self-supervision has been widely used in language modeling \cite{devlin2019bert,lan2019albert,liu2019roberta,baevski2020wav2vec,chen2020big,liu2018empower}, representation learning \cite{feng2019self,kolesnikov2019revisiting,sabokrou2019self,zhu2018spherical}, robotics \cite{mees2019self,nair2017combining,berscheid2020self}, computer vision \cite{zhai2019s4l,sun2019unsupervised,xu2019self,yin2020dreaming} and reinforcement learning \cite{kahn2018self,zeng2018learning,pong2019skew}. 
	One of our contributions is to generalize and extend the idea of self-supervision to the task of euphemism identification.


	\section{Problem Description}
	\label{sec:problem}
	
	In this study, we assume a content moderator has access to a textual corpus (\eg, a set of posts from an online forum), 
	and is required to moderate content related to a given list of target keywords. 
	In practice, forum users may use \emph{euphemisms}---words that are used as substitutes for  one of the target keywords. 
	We have two goals, euphemism detection and euphemism identification, defined as follows: 
	1) \emph{Euphemism detection:} Learn which words are being used as euphemisms for target keywords. A moderator can use this to filter content that may need to be moderated.
	2) \emph{Euphemism identification:} Learn the meaning of euphemisms. This can be used by the moderator to understand context, and individually review content that uses euphemisms. 
	
	As shown in Figure \ref{fig:model_overview}, these two tasks are complementary and form, together, a content moderation pipeline.
	The euphemism detection task takes as input (a) the raw text corpus, and (b) a list of target keywords (\eg, heroin, marijuana, ecstasy, \etc). 
	The expected output is an ordered ranked list of euphemism candidates, sorted by model confidence. 
	The euphemism identification module takes as input a euphemism (\eg, weed) and  outputs a probability distribution over the target keywords in the list. 
	For example, if we feed the euphemism \emph{weed} into this module, the output should be a probability distribution over keywords, with most of the mass on \emph{marijuana}.

	\noindent \textbf{Remark}: 
	We use the term ``category'' to denote a 
	topic (\ie, drug, weapon, sexuality). 
	We use ``target keyword'' to refer to the specific keyword in each category users might be trying to use euphemisms for (\eg, ``marijuana'' and ``heroin'' are examples of target keywords in the drug category).

	\section{Proposed Approach}
	\label{sec:model}
	We next discuss in detail our proposed euphemism detection approach in Section \ref{sec:model_det} and the proposed euphemism identification approach in Section \ref{sec:model_iden}.

	\subsection{Euphemism Detection}
	\label{sec:model_det}
	\begin{figure*}
		\centering
		\includegraphics[width=1.0\linewidth]{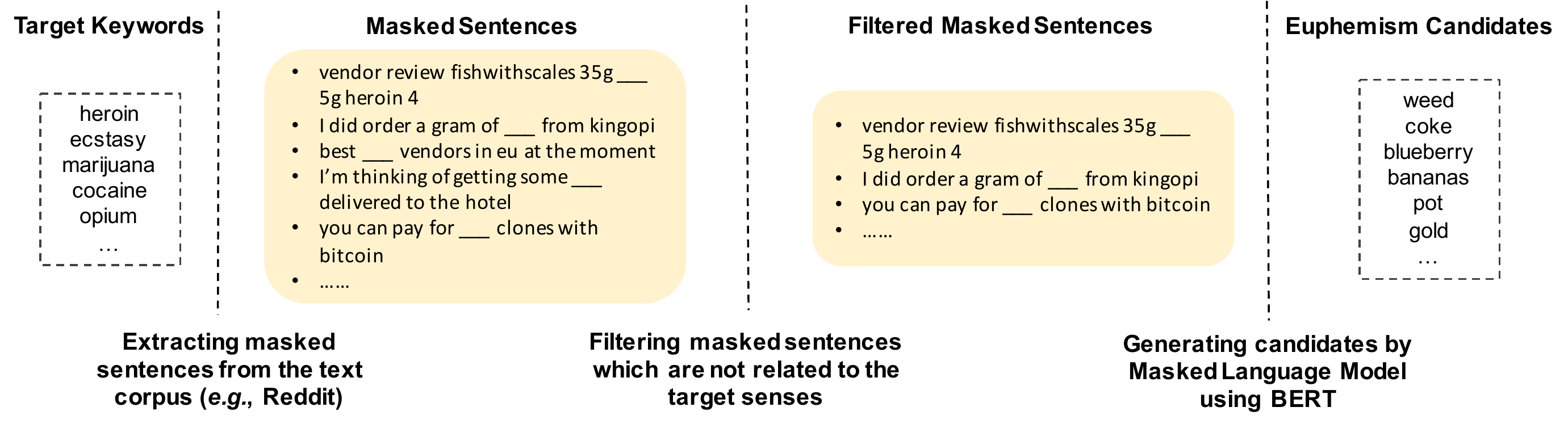}
		\caption{An overview of the euphemism detection framework.}
		\label{fig:model_det}
	\end{figure*}

	We formulate the euphemism detection problem as an unsupervised fill-in-the-mask problem and solve it by combining self-supervision with a Masked Language Model (MLM), an important modeling idea behind BERT \cite{devlin2019bert}. 
	Our proposed approach to euphemism detection has three stages (represented  in Figure~\ref{fig:model_det}): 
	1) Extracting contextual information, 
	2) Filtering out uninformative contexts, and 
	3) Generating euphemism candidates.  
	
	\noindent\textbf{Contextual information extraction.} Taking as input all the target keywords, this stage first extracts the masked sentences of all the keywords. 
	Here, a masked sentence refers to a sentence excluding the target keyword. 
	Taking the first example sentence in Table~\ref{table:example3} as an example, the corresponding masked sentence is ``This 22 year old former [MASK] addict who i did drugs with was caught this night.''. 
	A collection of all  masked sentences of the target keywords serves as the source of the relevant and crucial contextual information. 
	
	\noindent\textbf{Denoising contextual information.} Not all masked sentences are equally informative. 
	There may be  instances where the mask token (\ie, ``[MASK]'') can be filled by more than one target term, or words unrelated to the target terms, without affecting the quality of the sentence. 
	The fourth example sentence 
	in Table~\ref{table:example3} is one such case, where the masked sentence ``Why is it so hard to find [MASK]?'' is not specific to a drug; the mask token can be filled  by many words, including nouns such as ``jobs'', ``gold'' and even pronouns such as ``him''. 
	Such masked sentences (example sentences 4--6 
	in Table~\ref{table:example3}) are generic and lack relevant  context for disambiguating a polysemous word. 
	
	To filter such generic masked sentences, we propose a self-supervised approach that makes use of the Masked Language Model (MLM) proposed in BERT \cite{devlin2019bert}.
	Recall that self-supervision involves creating a new learning task from unlabeled data.
	An MLM aims to find suitable replacements of the masked token, and outputs a ranked list of potential replacement terms.  
	We start by fine-tuning the ``bert-base-uncased'' pre-trained model\footnote{ \url{https://huggingface.co/transformers/model\_doc/bert.html\#bertformaskedlm}} to the language of the  of domain-specific body of text (for instance, a collection of Reddit posts for identifying drug-related euphemisms).
	
	Empirically, we find that if a masked sentence is specific to a target category (\eg, drug names), words related to the target category will be ranked highly in the replacement list. 
	In contrast, if the masked sentence is generic, the highly ranked replacements are more likely  to be random words unrelated to the target category (\eg, ``jobs'', ``gold'', ``him''). 
	Therefore, we set an MLM threshold $t$ to filter out the generic masked sentences. 
	Considering the ranked list of replacements for the mask token, if any target keyword appears in the top $t$ replacement candidates for the masked sentence, we consider  the masked sentence to be a valid instance of a context. 
	Otherwise, it is considered to be a generic one and filtered out. 
	We set the threshold $t$ to $t=5$ in our experiments and discuss its  sensitivity in Section \ref{sec:dis_parameter_analysis}.

	\noindent\textbf{Candidate euphemism generation.} 
	Once we have (a) a pre-trained language model that is fine-tuned to the text corpus of interest, and (b) a filtered list of masked sentences, we are ready to  generate euphemism candidates. 
	For each masked sentence $m$, and for each word candidate $c$ in the vocabulary (\ie, all words available in the BERT pre-trained model), we compute its MLM probability (the probability of the word occurring in $m$ as predicted by the language model) $h_{c,m}$  by a pre-trained BERT model.
	Therefore, given a set of masked sentences, the weight $w_{c}$ of a word candidate $c$ is calculated as: 
	$w_c = \sum_{m'}h_{c, m'}$. 
	The final generation stage simply ranks all word candidates by their weights. 
	
	To clarify, we use the masked language model twice---once for filtering the masked sentences and a second time for generating the euphemism candidates from the masked sentences.

	\subsection{Euphemism Identification}
	\label{sec:model_iden}
	\begin{figure*}
		\centering
		\includegraphics[width=1.00\linewidth]{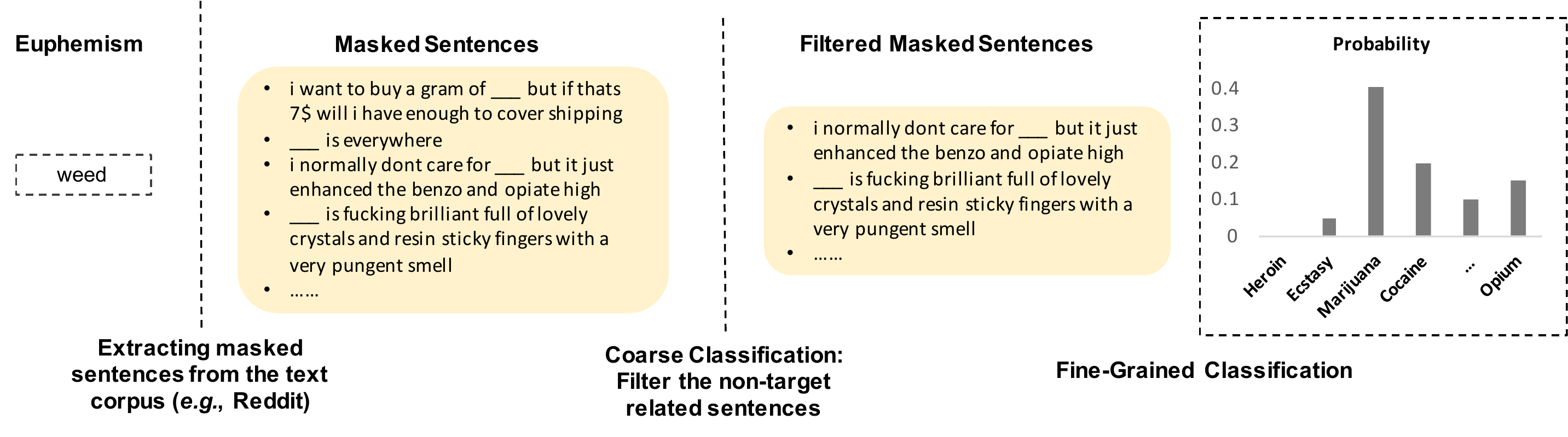}
		\caption{An overview of the euphemism identification framework.}
		\label{fig:model_iden}
	\end{figure*}
	
	Once the euphemisms are detected, we aim to identify what target keyword each euphemism refers to. 
	Taking the second and third example sentences in Table \ref{table:example1}, we want to identify that ``ice'' refers to ``methamphetamine'' and  ``pot''  to  ``marijuana''. 
	Euphemism identification has been acknowledged as a highly challenging task \cite{yuan2018reading}, due to two problems: 
	\begin{itemize}
		\item \textit{Resource challenge}: 
		No publicly-available, curated datasets are adequate to exhaustively learn a growing list of mappings between euphemisms and their target keywords. Moreover, it is unclear what linguistic and ontological resources one would need to automate this task.  
		\item \textit{Linguistic challenge}: 
		The distinction in meaning between the target keywords  (\eg, cocaine and marijuana) is often subtle and difficult to learn from raw text corpora alone. 
		Even human experts are often unable to accurately identify what a euphemism refers to by looking at a single  sentence. 
		A second linguistic challenge is related to the ambiguity of the euphemism itself. 
		A given euphemism can  be used in a euphemistic or non-euphemistic sense, adding the extra layer of linguistic nuance (Table \ref{table:example4}). 
	\end{itemize}

	
	
	We tackle the \textit{resource challenge} by designing a self-supervised learning scheme. 
	We extract all sentences that include the target keywords (\eg, cocaine, marijuana, heroin), mask the target keywords, and consider the masked sentences as training samples. 
	This allows us to automatically construct a labeled dataset, where the input samples are the masked sentences, and their respective target keywords are labels.
	
	To address the \textit{linguistic challenge}, we adopt a coarse-to-fine-grained classification scheme. 
	Such hierarchical schemes have shown better discriminative performance in various tasks \cite{huo2019coarse,liu2018global,li2019exploiting}. 
	The coarse classifier is a binary classifier that outputs whether a sentence is related to a specific category (\eg, drug) or not. 
	It aims to filter out sentences where the euphemism candidates do not occur in a euphemistic sense. 
	The fine-grained classifier is a multi-class classifier trained on the curated dataset from the self-supervised learning scheme; this aims to learn a specific mapping from the masked sentence to the target keyword. 
	We discuss the details of these classifiers below; first, we step through an example  of the end-to-end pipeline.
	
	\noindent \textbf{Example}: 
	Suppose our euphemism detection pipeline outputs the term ``weed''. 
	We aim to generate a probability distribution over target keywords, with most of the mass on marijuana (Figure \ref{fig:model_iden}). 
	Assume that we already have a trained coarse classifier and a trained fine-grained classifier (training details will be discussed below in \ref{sec:iden_approach}). 
	We first extract all  masked sentences that previously contained ``weed" from the text corpus. 
	Second, using the coarse classifier, 
	we filter out the masked sentences that are unrelated to the target category (\ie, all masked sentences that do not discuss something drug-related). 
	Then, we use the filtered masked sentences as inputs to the fine-grained multi-class classifier, and obtain the target keyword label for each masked sentence. 
	We now have a list of labels for the euphemism ``weed'' (\eg, 36,100 ``marijuana'' labels, 4,200 ``ecstasy'' labels, \etc) and the final output for a euphemism is a probability distribution by the number of labels for each target keyword.

	\subsubsection{Training Details}
	\label{sec:iden_approach}
	As discussed above, two classifiers need to be trained: 
	1) A coarse classifier to filter out the masked sentences of the euphemism words not associated with their euphemistic sense
	and, 2) A multi-class classifier to determine the target keyword to which the euphemism refers. 
	
	\noindent \textbf{Coarse Classifier}: 
	The coarse classifier is a binary classifier that decides whether a masked sentence is related to the target keywords or not. 
	Obtaining positive instances is easy: we collect all the masked sentences of the target keywords (\eg, we obtain the masked sentences from Table \ref{table:example3}). 
	To obtain the negative instances, we adopt a negative sampling approach \cite{mikolov2013distributed}; 
	we randomly choose a sentence in the whole 
	text corpus and randomly mask a token. 
	Since the corpus is large and diverse, we assume the randomly chosen masked sentence is unrelated to the target keyword. 
	With high probability, this assumption is correct.
	To create a balanced dataset, we select as many negative instances as there are positive ones. 
	This set of positive and negative instances constitutes the training set, with masked sentences and their respective labels to indicate whether a masked sentence is related to the target keywords or not. 
	We use 70\% of the data instances for training, 10\% for validation, and 20\% for testing. 
	We select an LSTM recurrent neural network model \cite{hochreiter1997long} with an attention mechanism \cite{bahdanau2015neural} for its ability to learn  to pay attention to the correct segments of an input sequence. 
	We obtain 98.8\% training accuracy 
	and 90.1\% testing accuracy. 
	Our experiments also include other classification models---we discuss our selection in Section \ref{sec:ablation_coarse}. 
	
	\noindent \textbf{Multi-class Classifier}:
	As presented above, we use as inputs the masked sentences and as labels the target keywords. 
	Empirically, we obtained good performance from a multinomial logistic regression classifier \cite{hosmer2013applied}.
	We first represent each word as a one-hot vector\footnote{One-hot encoding is used to represent a categorical variable whose values do not have an  ordinal relationship. 
		The one-hot encoding of a word $v_i\in V$, where $V$ denotes the vocabulary, is a $|V|$-dimensional vector of all zeros except for a 1 at the $i$th index.}.
	We then represented each sentence as the average of its member words' encodings. 
	By using the same data splitting ratio as the coarse classifier, we obtain a training accuracy of 55\% and a testing accuracy of 24\%  for the drug dataset (described in Section \ref{sec:res}). 
	As a point of comparison, with 33 target names in the drug dataset a random guess  would yield an accuracy of 3.3\%. 
	We discuss the results for other classification models in Section~ \ref{sec:ablation_fine-grained}.

	\section{Empirical Evaluation}
	\label{sec:res}
	In this section, we empirically evaluate the performance of our proposed approach and compare with that a set of baseline models on both euphemism detection (in Section \ref{sec:res_det}) and euphemism identification (in Section \ref{sec:res_iden}). 
	
	\subsection{Experimental Setup}
	We implemented all models in Python 3.7 and conducted all the experiments on a computer with twenty 2.9 GHz Intel Core i7 CPUs and one GeForce GTX 1080 Ti GPU. 
	
	\noindent \textbf{Datasets}: 
	We empirically validate our proposed model on three separate datasets related to three broad areas of euphemism usage: drugs, weapons, and sexuality. 
	For the algorithm to be applicable to a dataset, we require two kinds of inputs: 1) the raw text corpus from which we extract the euphemisms and their masked sentences, and 2) a list of target keywords (\eg, heroin, marijuana, ecstasy, \etc). 
	For the purpose of carrying out a quantitative evaluation of the euphemism detection and identification approaches and comparing them with prior art, we rely on a ground truth list of euphemisms and their target keywords. Ideally, such a list  should contain all euphemisms for the evaluation of euphemism detection, and a one-to-one mapping from each euphemism to its actual meaning, for the evaluation of euphemism identification. 
	
	\begin{itemize}
		\item {\em Drug dataset}: 
		From a publicly available data repository \cite{redditcorpus},
		we extracted 1,271,907 posts from 46 distinct 
		``subreddits''\footnote{Forums hosted on the Reddit website, 
			and associated with a specific topic.} 
		related to drugs and dark web markets, 
		including the largest ones---``Bitcoin'' (565,614 posts), 
		``Drugs'' (373,465 posts),
		``DarkNetMarkets'' (125,300 posts),
		``SilkRoad'' (22,989 posts), 
		``DarkNetMarketsNoobs'' (22,699 posts).
		A number of these subreddits were banned from the platform 
		in early 2018 \cite{cimpanu2018reddit}. 
		As a result, the posts collected were authored between February 9, 2008 and December 31, 2017. 
		While online drug trade dates back (at least) to USENET groups in the 1990s,
		it truly picked up mainstream traction with the emergence of the 
		Silk Road in 2011. 
		Our data corpus captures these early days, 
		as well as the more mature ecosystem that followed~\cite{soska15markets}.
		
		For ground truth, we use a list of drug names and corresponding euphemisms
		compiled by the (USA) Drug Enforcement Administration~\cite{drug2018slang}.
		This list is intended as a practical reference for law enforcement personnel.
		Due to the rapidly evolving language used in the drug-use subculture,
		it cannot be comprehensive or error-free,
		but it is the most reliable ground truth available to us.
		
		\item {\em Weapon dataset}: The raw text corpus comes from a combination of the 
		corpora collected by Zanettou et al.\ \cite{zannettou2018gab}, Durrett et al.\ \cite{durrett2017identifying}, Portnoff et al.\  \cite{portnoff2017tools} and the examples in Slangpedia\footnote{\url{https://slangpedia.org/}}. 
		The combined corpus has 310,898 posts. 
		Both the ground truth list of weapon target keywords and the respective euphemisms are obtained from The Online Slang Dictionary\footnote{\url{http://onlineslangdictionary.com/}} (one of the most comprehensive slang thesaurus available), Slangpedia, and The Urban Thesaurus\footnote{\url{https://urbanthesaurus.org/}}.
		
		\item {\em Sexuality dataset}: The raw text corpus comes from the Gab social networking services\footnote{\url{https://gab.com/}}. We use 2,894,869 processed posts, collected from Jan 2018 to Oct 2018 by PushShift.\footnote{Available at \url{https://files.pushshift.io/gab/}} Both the ground truth list of sexuality target keywords and the euphemisms are obtained from The Online Slang Dictionary. 
	\end{itemize}

	\subsection{Euphemism Detection}
	\label{sec:res_det}
	We evaluate the performance of euphemism detection in this section. 
	
	\noindent \textbf{Evaluation Metric}: 
	For each dataset, the input is an unordered list of target keywords and the output is an ordered ranked list of euphemism candidates. 
	Given the nature of the output, we evaluate the output using the metric precision at $k$ ($P@k$), which is commonly used in information retrieval to evaluate how well the search results corresponded to a query \cite{manning2008introduction}. 
	$P@k$, ranging from 0 to 1, measures the proportion of the top $k$ generated results that are correct (in our case, valid euphemisms), which we calculate with respect to the ground truth list for each dataset. 
	In cases where an  algorithm recovers only one word of a multi-word euphemism (e.g., ``Chinese" instead of ``Chinese tobacco"), we treat the candidate as incorrect.
	Because of the known shortcoming that $P@k$ fails to take into account the positions of the relevant documents \cite{jarvelin2017ir}, we report $P@k$ for multiple values of $k$ ($k=10, 20, 30, 40, 50, 60, 80, 100$) to resolve the issue. 
	
	We are unable to measure recall for the following two reasons: 
	1) Some euphemisms in the ground truth list do not appear in the text corpus at all and using recall as a measure can result in a misrepresentation of the performance of the approaches; 
	2) Those euphemisms that indeed appear in the text corpus,  may not have been used in the euphemistic sense. 
	For example, ``chicken" is a euphemism for ``methamphetamine,'' but it could have been used only in the animal sense in the corpus. 
	

	\begin{table*}[ht!]
		\centering
		\small
		\caption{Results on euphemism detection. Best results are in bold.}
		\begin{tabular}{c|c|cccccccc}
			\toprule
			\multicolumn{2}{c}{}& \textbf{$P@10$}  & \textbf{$P@20$} &  \textbf{$P@30$} &  \textbf{$P@40$} & \textbf{$P@50$}  & \textbf{$P@60$} &  \textbf{$P@80$} &  \textbf{$P@100$}\\
			\midrule
			
			\multirow{8}{*}{\rotatebox[origin=c]{90}{\textbf{Drug}}}
			&\textbf{Word2vec} & 0.10 & 0.10 & 0.09 & 0.09 & 0.08 & 0.09 & 0.08 & 0.09 \\
			&\textbf{TF-IDF + word2vec} & 0.30 & 0.25 & 0.20 & 0.20 & 0.16 & 0.17 & 0.16 & 0.18 \\
			&\textbf{CantReader \cite{yuan2018reading}} & 0.00 & 0.00 & 0.07 & 0.10 & 0.08 & 0.12 & 0.12 & 0.10 \\
			&\textbf{SentEuph \cite{felt2020recognizing}} & 0.10 & 0.10 & 0.07 & 0.05 & 0.08 & 0.07 & 0.09 & 0.07 \\
			&\textbf{EigenEuph \cite{magu2018determining}} & 0.30 & 0.30 & 0.30 & 0.25 & 0.22 & 0.22 & 0.20 & 0.19 \\
			&\textbf{GraphEuph \cite{taylor2017surfacing}} & 0.20 & 0.15 & 0.13 & 0.13 & 0.14 & 0.17 & 0.14 & 0.11 \\
			&\textbf{MLM-no-filtering} & 0.30 & 0.30 & 0.28 & 0.30 & 0.26 & 0.26 & 0.28 & 0.26 \\
			&\textbf{Our Approach} & \textbf{0.50} & \textbf{0.45} & \textbf{0.47} & \textbf{0.42} & \textbf{0.46} & \textbf{0.42} & \textbf{0.38} & \textbf{0.36} \\
			\midrule
			
			\multirow{8}{*}{\rotatebox[origin=c]{90}{\textbf{Weapon}}}
			&\textbf{Word2vec} & 0.30 & 0.30 & 0.27 & 0.23 & 0.18 & 0.20 & 0.20 &  0.18\\
			&\textbf{TF-IDF + word2vec} & 0.30 & 0.25 & 0.20 & 0.17 &  0.16 & 0.18 & 0.20 & 0.18 \\
			&\textbf{CantReader \cite{yuan2018reading}} & 0.20 & 0.20 & 0.17 & 0.18 & 0.16 & 0.17&0.13 & 0.11\\
			&\textbf{SentEuph \cite{felt2020recognizing}} & 0.00 & 0.00 & 0.03 & 0.05 & 0.06 & 0.05 & 0.05 & 0.04\\
			&\textbf{EigenEuph \cite{magu2018determining}} & 0.30 & 0.20  & 0.13 & 0.10 & 0.08 & 0.07 & 0.05 & 0.04\\
			&\textbf{GraphEuph \cite{taylor2017surfacing}} &  0.00 & 0.05 & 0.03 & 0.05 & 0.04 & 0.03 & 0.03 & 0.02\\
			&\textbf{MLM-no-filtering} & 0.30 & 0.30 & 0.20 & 0.17 & 0.18 & 0.18 & 0.15 & 0.15 \\
			&\textbf{Our Approach} & \textbf{0.40} & \textbf{0.45} & \textbf{0.37} & \textbf{0.35} & \textbf{0.32} & \textbf{0.28} & \textbf{0.25} & \textbf{0.20} \\
			\midrule
			
			\multirow{8}{*}{\rotatebox[origin=c]{90}{\textbf{Sexuality}}}
			&\textbf{Word2vec} & 0.10 & 0.05 & 0.07 & 0.08 & 0.08 & 0.08 & 0.09 &  0.09 \\
			&\textbf{TF-IDF + word2vec} & 0.40 & 0.25 & 0.20 & 0.20 & 0.20 & 0.17 & 0.15 & 0.13  \\
			&\textbf{CantReader \cite{yuan2018reading}} & 0.10 & 0.10 & 0.07 & 0.08 & 0.06 & 0.08 & 0.09 & 0.10 \\
			&\textbf{SentEuph \cite{felt2020recognizing}} & 0.10 & 0.10 & 0.08 & 0.10 & 0.08 & 0.10 & 0.08 & 0.06\\
			&\textbf{EigenEuph \cite{magu2018determining}} & 0.20 & 0.15 & 0.13 & 0.15 & 0.16 & 0.18 & 0.14 & 0.11\\
			&\textbf{GraphEuph \cite{taylor2017surfacing}} & 0.00 & 0.00 & 0.03 & 0.05 & 0.04 & 0.03 & 0.04 & 0.03 \\
			&\textbf{MLM-no-filtering} & 0.50 & \textbf{0.40} & 0.30 & 0.23 & 0.22 & 0.22 & 0.19 & 0.15 \\
			&\textbf{Our Approach} & \textbf{0.70} & \textbf{0.40}& \textbf{0.33}& \textbf{0.33}& \textbf{0.28}& \textbf{0.25}& \textbf{0.23}& \textbf{0.19} \\
			\bottomrule
		\end{tabular}
		\label{table:res_dec}
	\end{table*}
	
	\noindent \textbf{Baselines}: 
	We compare our proposed approach with the following competitive baseline models:
	
	\begin{itemize}
		\item \textbf{Word2vec}: We use the word2vec algorithm \cite{mikolov2013distributed,mikolov2013efficient}  to learn the word embeddings (100-dimensional) for all the words separately for the Drug, Weapon and Sexuality datasets. We then choose as euphemism candidates those words that are most similar to the input target keywords, in terms of cosine similarity (average similarity between the word and all input target keywords). 
		This approach relates words by implicitly accounting for the context in which they occur. 
		\item \textbf{TF-IDF + word2vec}: Instead of treating all the words in the dataset equally, this method first ranks the words by their potential to be euphemisms. Toward this, we calculate the TF-IDF weights of the words \cite{manning2008introduction} with respect to a background corpus (\ie, Wikipedia\footnote{\url{https://dumps.wikimedia.org/enwiki/}}), which captures a combination of the frequency of a word and its distinct usage in a given corpus. The idea is inspired by the assumption that words ranked higher based on TF-IDF in the target corpus have a greater chance of being euphemisms than those ranked lower \cite{magu2018determining}.  After the pre-selection by TF-IDF, we then generate the euphemism candidates by following the Word2vec approach above.  
		\item \textbf{CantReader}\footnote{\url{https://sites.google.com/view/cantreader}} \cite{yuan2018reading} employs a neural-network based embedding technique to analyze the semantics of words, detecting the euphemism candidates whose contexts in the background corpus (\eg, Wikipedia) are significantly different from those in the dark corpus. 
		\item \textbf{SentEuph} \cite{felt2020recognizing} recognizes euphemisms by the use of sentiment analysis. It lists a set of euphemism candidates using a bootstrapping algorithm for semantic lexicon induction. For a fair comparison with our approach, we do not include the manual filtering stage of the algorithm proposed by Felt and Riloff \cite{felt2020recognizing}. 
		\item \textbf{EigenEuph} \cite{magu2018determining} leverages word embeddings and a community detection algorithm, to generate a cluster of euphemisms by the ranking metric of eigenvector centralities. 
		\item \textbf{GraphEuph}\footnote{\url{https://github.com/JherezTaylor/hatespeech_codewords}} \cite{taylor2017surfacing} also identifies euphemisms using word embeddings and a community detection algorithm. Specifically, it creates neural embedding models that capture word similarities, uses graph expansion and the PageRank scores \cite{page1999pagerank} to bootstrap an initial set of seed words, and finally enriches the bootstrapped words to learn out-of-dictionary terms that behave like euphemisms. 
		\item \textbf{MLM-no-filtering} is a simpler version of our 
		proposed approach and shares its architecture. The key difference from our proposed approach is that instead of filtering the noisy masked sentences, it uses them \textit{all} to generate the euphemism candidates. In effect, this baseline serves as an ablation to understand the effect of the filtering stage. 
	\end{itemize}
	
	For a fair comparison of the baselines, we experimented with different combinations of parameters and report the best performance for each baseline method. 
	
	\noindent \textbf{Results}: 
	Table \ref{table:res_dec} summarizes the euphemism detection results. 
	Our proposed approach outperforms all the baselines by a wide margin for the different settings of the  evaluation measure on all the three datasets we studied. 
	
	The most robust baselines over the three datasets are TF-IDF + word2vec, EigenEuph  and MLM-no-filtering. 
	When compared with Word2vec, the superior performance of TF-IDF + word2vec lies in its ability to select a set of potential euphemisms  by calculating the TF-IDF with a background corpus (\ie, Wikipedia). 
	While this pre-selection step works well (relative to Word2vec) on the Drug and Sexuality datasets, it does not impact the performance on the Weapon dataset. A plausible explanation for this is that the euphemisms do not occur very frequently in comparison with the other words in the Weapons corpus and therefore, are not ranked highly by the TF-IDF scores.

	SentEuph \cite{felt2020recognizing}'s comparatively poor performance 
	is explained by the absence of the required additional manual filtering stage to refine the results. 
	As mentioned before, this was done to compare the approaches based on their automatic performance alone.
	GraphEuph \cite{taylor2017surfacing} shows a reasonable performance on the Drug dataset, but fails to detect weapon- and sexuality-related euphemisms. This limits the generalization of the approach that was tested only on a hate speech dataset by Taylor \etal \cite{taylor2017surfacing}. 
	The approach of CantReader \cite{yuan2018reading} seems to be ineffective because not only does it require additional corpora to make semantic comparisons---a requirement that is ill-defined because the nature of the additional corpora needed for a given dataset is not specified---but also because the results of CantReader are quite sensitive to parameter tuning. We were unable to reproduce the competitive results reported by Yuan \etal \cite{yuan2018reading}, even after multiple  personal communication attempts with the authors. 
	By comparing the performance of our approach and that of the ablation MLM-no-filtering, we conclude that the proposed filtering step is effective in eliminating the noisy masked sentences and is indispensable for reliable results.

	\noindent \textbf{False positive analysis}: 
	By studying the false positives in our results, we recovered several euphemisms that were not included in our ground truth list. 
	Table~\ref{fig:casestudies-detection2} in Appendix \ref{sec:appendix} shows sentences associated with 10 of the top 16 false positive euphemisms from the drug dataset. 
	Several of these are true euphemisms for drug keywords that were not present in the DEA ground truth list (\eg, md, l, mushrooms).
	Others are not illicit drugs (\eg, alcohol, cigarettes), but they are used in this corpus in a way that is closely related to how people use drug names,  and reveal new usage patterns. 
	For example, the sentences for ``cigarettes" indicate that people appear to be combining cigarette use with other drugs, such as PCP. 
	Similarly, the sentences containing ``alcohol" reveal that people are dissolving illicit drugs in alcohol. 
	Of these 10 false positives (according to our ground truth dataset), only five are actually false positives; these words are semantically related to the drug keywords, but they are not proper euphemisms (\eg, ``pressed" is a form factor for drug pills).

	\subsection{Euphemism Identification}
	\label{sec:res_iden}
	For each euphemism that we have successfully detected, we now evaluate euphemism identification. 
	
	\noindent \textbf{Evaluation Metric}: 
	For each euphemism, we generate a probability distribution over all target keywords and therefore, obtain a ranked list of the target keywords. 
	We evaluate the top-$k$ accuracy ($Acc@k$), which measures how often the ground truth label (target keyword) falls in the top $k$ values of our generated ranked list.

	\noindent \textbf{Baselines}: 
	Given the lack of related prior work for the task of euphemism identification, we establish a few baseline methods and compare our proposed approach with them.
	
	\begin{itemize}
		\item \textbf{Word2vec}: 
		For each euphemism, we select the target keyword that is closest to it using the measure of cosine similarity. Here we compare the word embeddings (100-dimensional) obtained by training the word2vec algorithm \cite{mikolov2013distributed,mikolov2013efficient}  on each text corpus separately. 
		\item \textbf{Clustering + word2vec}: For each euphemism, we cluster all its masked sentences, represented as the average of the word embeddings of the component words, using a $k$-means algorithm (we set $k=2$). By clustering, our aim is to separate the masked sentences into two groups (ideally one group of informative masked sentences and the other group of uninformative masked sentences as presented in Table \ref{table:example3}) and to filter out the uninformative masked sentences that are not related to the target keywords. Then, we compare the embeddings of the filtered masked sentences of the euphemism and the target keywords using the measure of cosine similarity. The target keyword that is most similar to the filtered masked sentences is selected for identification. 
		\item \textbf{Binary + word2vec}: similar to our approach, we use a binary classifier to filter out noisy masked sentences that are not related to the target keywords. Then, we use the Word2vec approach above to find its closest target keyword. 
		\item \textbf{Fine-grained-only} is an simplistic version of our approach, which only uses the fine-grained multi-class classifier, without the preceding coarse classifier. 
	\end{itemize}

	\noindent \textbf{Results}: 
	Table \ref{table:res_iden} summarizes the euphemism identification results. 
	There are 33, 9, and 12 categories for the drug, weapon and sexuality datasets respectively, resulting in a random guess performance for $Acc@1$ to be 0.03, 0.11, 0.08 (\ie the inverse of the number of categories). 
	Our algorithm achieves the best performance for all three datasets and has a large margin over the random guess performance.

	Word2vec exhibits poor performance, in that it is unable to capture the nuanced differences between the target keywords by taking all sentences into consideration. 
	Therefore, we construct two baselines (\ie, Clustering + word2vec and Binary + word2vec) to remove the noisy sentences and aid learning using a more homogeneous set of masked sentences. 
	Empirically, we find that a binary classifier contributes more towards the performance, compared to the clustering algorithm. 
	This is because, the result of clustering 
	did not adequately cluster the sentences into a target keyword cluster and a non-target keyword cluster. 
	Taking the drug dataset as an example, 
	we found that owing to the widely varying contexts and vocabulary diversity of the dataset,
	the clustering results were inadequate. 
	For instance, a qualitative examination of the results of clustering for a few euphemisms showed that the cluster separation sometimes occurred by the ``quality'' attribute (\eg, high quality vs. low quality drugs) or even sentiment (\eg, feeling high vs. feeling low). 
	Therefore, $k$-means clustering fails as a filter for the non-drug-related masked sentences and does not lead to performance improvement. We leave exploring other clustering algorithms for future work.
	In contrast, the binary classifier, which can be taken as a directed $k$-means clustering algorithm, specifically filters out the non-drug-related sentences and is therefore a helpful addition. 
	For such a specific task, the binary classifier performance can be taken as a performance upper bound for clustering algorithms. 
	
	We highlight two important findings: 
	1) By comparing the results of Word2vec and Fine-grained-only, 
	we demonstrate the advantage of using a classification algorithm over an unsupervised word embedding-based method; 
	2) By comparing the differences between Word2vec and Binary + word2vec, and the differences between Fine-grained-only and our approach, we demonstrate the superior discriminative ability of a binary filtering classifier and therefore, highlight the benefit of using a coarse-to-fine-grained classification over performing only multi-class classification.

	\begin{table}[ht!]
		\centering
		\small
		\caption{Results on euphemism identification. Best results are in bold.}
		\begin{tabular}{c|c|ccc}
			\toprule
			\multicolumn{2}{c}{}& \textbf{$Acc@1$}  & \textbf{$Acc@2$} &  \textbf{$Acc@3$} \\
			\midrule
			
			\multirow{5}{*}{\rotatebox[origin=c]{90}{\textbf{Drug}}}
			&\textbf{Word2vec} & 0.07 & 0.14 & 0.21 \\
			&\textbf{Clustering + word2vec} & 0.06 & 0.15 & 0.25 \\
			&\textbf{Binary + word2vec} & 0.13 & 0.22 & 0.30 \\
			&\textbf{Fine-grained-only} & 0.11 & 0.19 & 0.26 \\
			&\textbf{Our Approach} & \textbf{0.20} & \textbf{0.31} & \textbf{0.38} \\
			\midrule		
			
			\multirow{5}{*}{\rotatebox[origin=c]{90}{\textbf{Weapon}}}
			&\textbf{Word2vec} & 0.10 & 0.27 &  0.40 \\
			&\textbf{Clustering + word2vec} & 0.11 & 0.25 & 0.37 \\
			&\textbf{Binary + word2vec} & 0.22 & 0.43 & 0.57 \\
			&\textbf{Fine-grained-only} & 0.25  & 0.40 & 0.61 \\
			&\textbf{Our Approach} & \textbf{0.33} & \textbf{0.51} & \textbf{0.67} \\
			\midrule
			
			\multirow{5}{*}{\rotatebox[origin=c]{90}{\textbf{Sexuality}}}
			&\textbf{Word2vec} & 0.17 & 0.22 & 0.42 \\
			&\textbf{Clustering + word2vec} & 0.15 & 0.30 & 0.49 \\
			&\textbf{Binary + word2vec} & 0.21 & 0.39 & 0.59 \\
			&\textbf{Fine-grained-only} &  0.19 & 0.40 & 0.51 \\
			&\textbf{Our Approach} & \textbf{0.32} & \textbf{0.55} & \textbf{0.64} \\
			\bottomrule
		\end{tabular}
		\label{table:res_iden}
	\end{table}

	\section{Discussion}
	\label{sec:dis}
	Our algorithms rely on a relatively small number of hyper-parameters and choices of classification models. 
	In this  section, we demonstrate how to choose these hyper-parameters through detailed ablation studies, primarily on  the drug dataset.
	
	\subsection{Ablation Studies for Euphemism Identification}
	As discussed above, we adopt a coarse-to-fine-grained classification scheme for euphemism identification, relying on two classifiers used in cascade. 
	We discuss here the performance of multiple classifiers on both coarse and fine-grained classification. 
	
	\subsubsection{Coarse Classifiers}
	\label{sec:ablation_coarse}
	\begin{figure}
		\centering
		\includegraphics[width=0.7\linewidth]{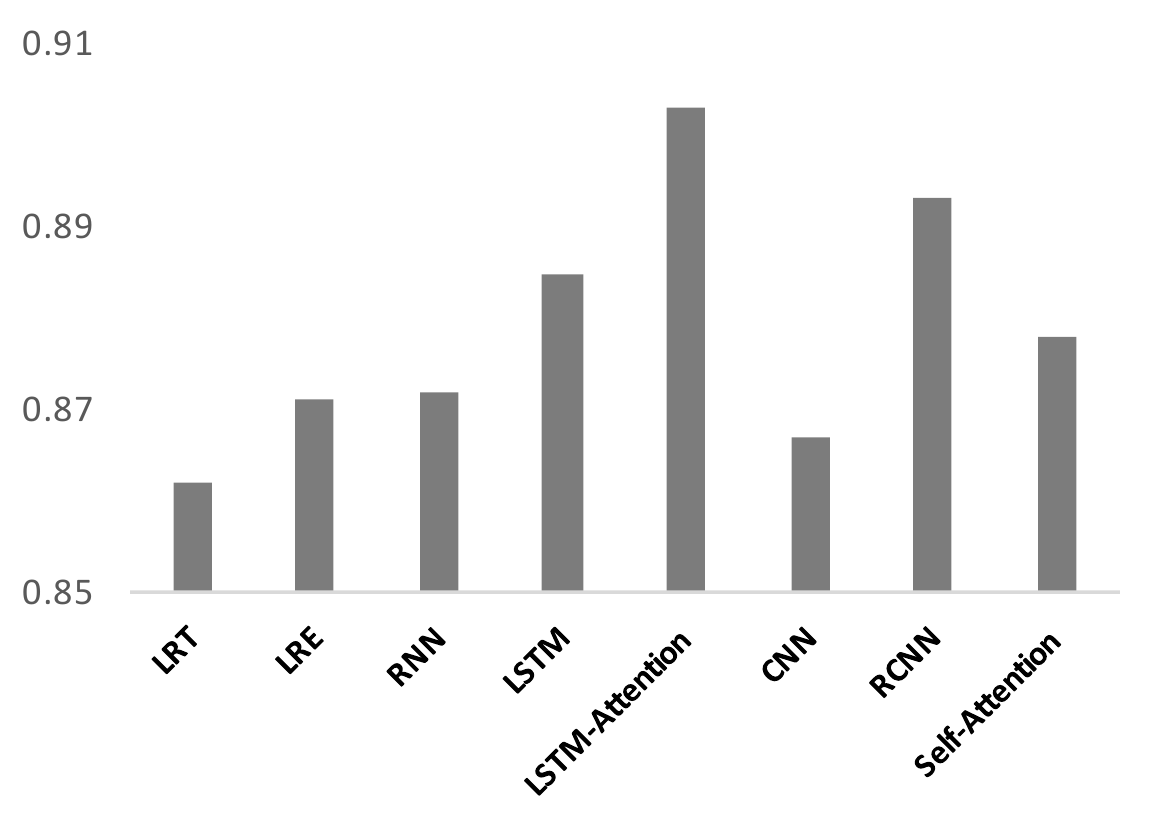}
		\caption{Testing accuracy for the coarse classifier.}
		\label{fig:11}
	\end{figure}
	
	In the euphemism identification framework, we use a binary classifier to filter out the sentences where euphemisms are used in non-euphemistic senses. 
	We experiment with the binary classifiers shown below. Note that for all the neural models, we use 100-dimensional GloVe embeddings\footnote{\url{https://nlp.stanford.edu/projects/glove/}} \cite{pennington2014glove} pre-trained on Wikipedia and tune the embeddings by the models. 
	\begin{itemize}
		\item Logistic Regression \cite{hosmer2013applied} on raw Text (LRT): we first represent each word as a one-hot vector and then represent each sentence as the average of its member words' encodings. 
		\item Logistic Regression on text Embeddings (LTE): we learn the word embeddings (100-dimensional) using word2vec \cite{mikolov2013distributed,mikolov2013efficient}. 
		We represent each sentence by the average of its member words' embeddings. 
		\item Recurrent Neural Network (RNN) \cite{rumelhart1985learning}: we use a 1-layer bidirectional RNN with 256 hidden nodes. 
		\item Long Short-Term Memory (LSTM) \cite{hochreiter1997long}: we use a 1-layer bidirectional LSTM with 256 hidden nodes. 
		\item LSTM-Attention: we add an attention mechanism \cite{bahdanau2015neural} on LSTM. 
		\item Convolutional Neural Networks (CNN) \cite{kim2014convolutional}: we train a simple CNN with one layer of convolution on top of word embeddings. 
		\item Recurrent Convolutional Neural Networks (RCNN) \cite{lai2015recurrent}: we apply a bidirectional LSTM and employ a max-pooling layer across all sequences of texts. 
		\item Self-Attention \cite{lin2017structured}: instead of using a vector, we use a 2-D matrix to represent the embedding, with each row of the matrix attending on a different part of the sentence. 
	\end{itemize}

	We split the datasets into 70-10-20 for training, validation and testing. 
	The model parameters are tuned on the validation data. 
	Empirically, we find the LSTM-Attention performs the best across three datasets. This is  why we ultimately selected it and reported results using it in Section~\ref{sec:res}. 
	Yet, as shown in Figure \ref{fig:11}, other classifiers have satisfactory performance as well, and reach a testing accuracy ranging from 0.86 to 0.90.

	\subsubsection{Fine-Grained Classifiers}
	\label{sec:ablation_fine-grained}
	In the euphemism identification framework, we use a multi-class classifier to identify to which target keyword each euphemism refers. 
	Again, we experimented with the same set of classifiers as above.
	Interestingly, we find that, for fine-grained classification, 
	all classifiers have highly similar results. 
	One possible reason is that each class has relatively small number of training instances (ranging from a few hundreds to 100k), which limits the discriminative power of advanced algorithms. 
	For the drug dataset (33 target keywords), the training accuracy is about 55\% and the testing accuracy is about 24\%. 
	This shows the feasibility of the task since the random guess accuracy would be 3.3\%. 
	Given the similar performance across  classifiers, we recommend Logistic Regression on raw Text (LRT) for better computational efficiency.

	For both coarse classifiers and fine-grained classifiers, we leave more advanced classification algorithms for future work.

	\subsection{Parameter Analysis}
	\label{sec:dis_parameter_analysis}
	\begin{figure}[ht!]
		\centering
		\vspace{-0.3cm}
		\includegraphics[width=0.7\linewidth]{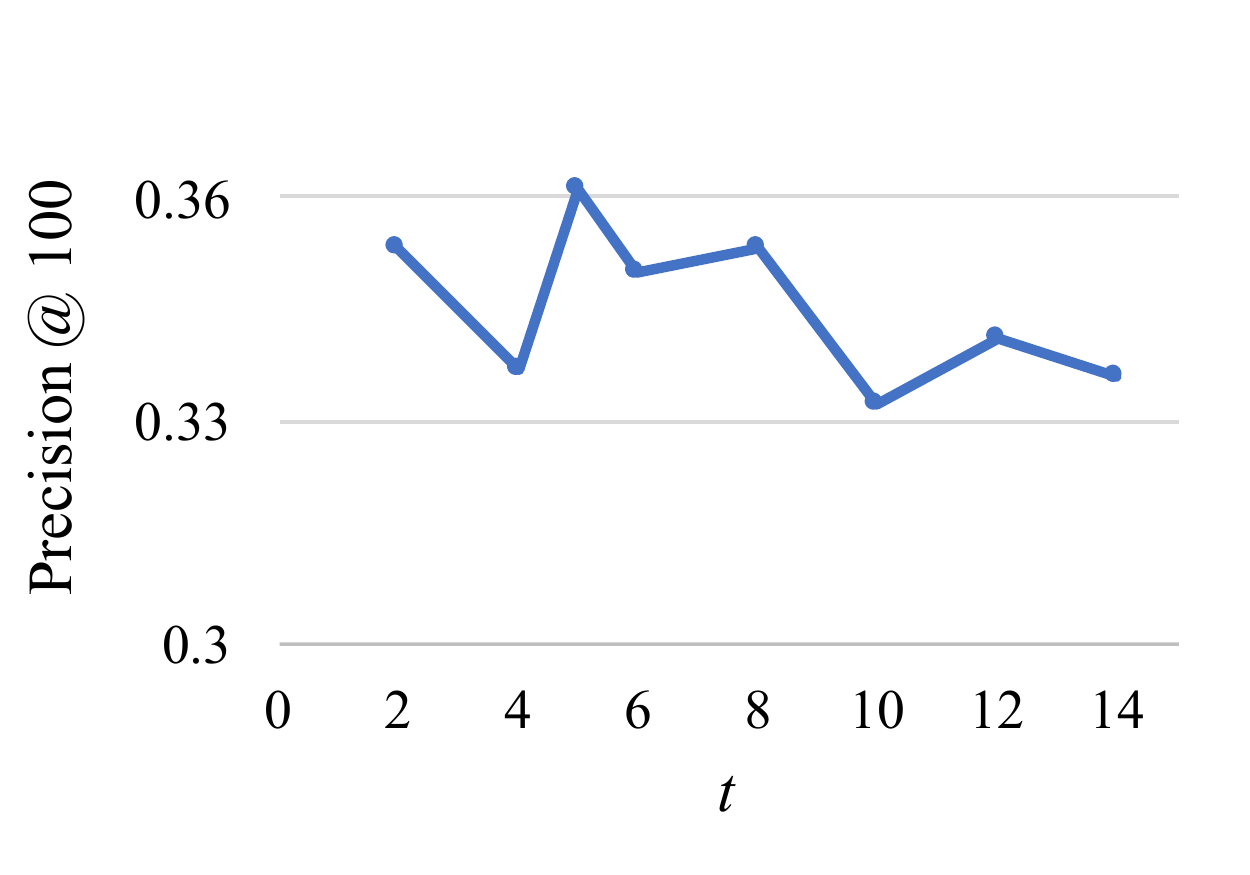}
		\caption{Sensitivity of $t$.}
		\label{fig:12}
	\end{figure}
	
	In the euphemism detection step 
	(Section \ref{sec:model_det}), 
	we set a masked language model threshold~$t$ to filter out the generic masked sentences. 
	In the ranked list of replacements for the mask token, if any target keyword appears in the top-$t$ 
	replacement candidates for the masked sentence, 
	we consider  the masked sentence a valid context instance. 
	Otherwise, we consider the masked sentence generic and filter it out. 
	Figure \ref{fig:12} shows how the results change with the threshold $t$ and we observe a slight decrease when the threshold $t$ is larger than 5. Therefore, $t=5$ appears to be an optimal parameter choice.


	\subsection{Limitations} 
	While our approach for euphemism detection and identification
	appears highly promising,
	it does have some limitations.
	
	\medskip 
	\noindent{\textbf{Text-only moderation}}: 
	Our approach only works with text, and our techniques
	are not easily generalizable to other media.
	Social media posts frequently include images, video, and audio,
	which can be even more challenging (and even more traumatic)
	to moderate by hand~\cite{TheVerge:Mods, TheVerge:Mods2, Ofcom:AI2019}.
	However, text is frequently associated with these other media, 
	\eg, in the form of comments, and thus detecting euphemism use might 
	indirectly provide clues to content moderators dealing with 
	different media.

	\medskip
	\noindent{\textbf{Other contexts}}:
	Our approach performs well
	on corpora discussing drugs, weapons, and sexuality.
	In preliminary experiments
	with a corpus of hate speech
	it did not perform nearly as well,
	producing many false matches
	when tasked with identifying racial slurs.
	We believe this is because
	euphemisms related to drugs, weapons, and sex
	typically have specific meanings;
	\eg, ``pot'' always refers to marijuana, not some other drugs.
	Racial slurs, on the other hand,
	are (in this corpus)
	used imprecisely, and interchangeably with generic swearwords,
	which seems to confuse euphemism detection.
	We do not know yet whether this is a fundamental limitation. 
	Even if it is, though,
	there are many  contexts where euphemisms have specific meanings
	and our approach should be effective, particularly forums selling illicit goods. 

	\medskip
	\noindent{\textbf{Robustness to adversarial evasion}}: 
	In our evaluation, we have relied on \textit{a priori} 
	non-adversarial datasets, 
	that were gleaned from public, online forums. 
	In other words, people were using euphemisms, 
	but we do not know whether they were using them
	specifically to evade content moderation.
	Perhaps these euphemisms are, for them,
	simply the ordinary names of certain things
	within the circle where they were discussing them.
	(Someone who consistently spoke of ``marijuana'' instead of ``pot''
	on a forum dedicated to discussing drug experiences
	might well be suspected of being an undercover cop.)
	
	Because our algorithms rely on sentence-level context
	to detect and identify euphemisms,
	an adversary would need to change that context
	to escape detection.
	Such changes 
	may also render the text unintelligible
	to its intended audience.
	Therefore, we expect our techniques
	to be moderately resilient to adversarial evasion.
	However, we cannot test our expectations at the moment,
	since we do not have a dataset
	where people were purposely using euphemisms
	\emph{only} to escape detection.
	
	\medskip
	\noindent{\textbf{Usability for content moderators}}:
	While our approach shows encouraging performance in lab tests,
	we have not yet evaluated
	whether it is good enough to be helpful to content moderators in practice.
	That evaluation would require a user study of professional content moderators.
	This is out of scope for the present paper,
	which  focuses on the technical underpinnings
	of euphemism detection and identification.
	We are interested in investigating usability as a follow-up study.
	
	As a preliminary experiment, we investigated the Perspective API\footnote{\url{https://www.perspectiveapi.com/}}, Google's automated toxicity detector to identify the likelihood of a sentence being considered toxic by a reader. 
	Perspective is reportedly used today by human moderators to filter or prioritize comments that may require moderation. 
	We take sentences from our datasets that contain the target keywords (\eg, ``marijuana'', ``heroin'')
	and for each such sentence, we evaluate the toxicity score of the sentence (a) with the target keyword, and (b) by replacing the target keyword 
	with one of its identified euphemisms (\eg, ``weed'', ``dope'').
	By comparing the toxicity scores, we can estimate the likelihood that a human moderator who is using Perspective API would be shown each version of the sentence. 
	Table \ref{table:discussion_perspective} shows the average toxicity scores when comparing 1000 randomly chosen original sentences with their euphemistic replacements for the drug, weapon, and sexuality categories. 
	We observe that sentences with target keywords have higher (or at least comparable) toxicity scores compared to sentences with euphemisms, which suggests that euphemisms could help escape content moderation based on the Perspective API. 
	In turn, detecting and identifying euphemisms could help defeat such evasive techniques. 
	
	\begin{table}[ht]
		\centering
		\small
		\caption{Average toxicity socres by Perspective API. (A): original sentences; (B): sentences with their euphemistic replacements.}
		\begin{tabular}{cccc}
			\toprule
			\multicolumn{1}{c}{}& \textbf{Drug}  & \textbf{Weapon} &  \textbf{Sexuality} \\
			\midrule		
			\textbf{A} & 0.209 & 0.235 & 0.612 \\	
			\textbf{B} & 0.178 & 0.232 & 0.522 \\	
			\bottomrule
		\end{tabular}
		\label{table:discussion_perspective}
	\end{table}



	\subsection{Ethics} 
	This study relies extensively on user-generated content. We consider here 
	the ethical implications of this work.  The data we use in this paper were posted on publicly accessible websites, and 
	do not contain any personal identifiable information (i.e., 
	no real names, email 
	addresses, IP addresses, etc.). Further, they 
	are from 2018 or earlier, 
	which greatly reduces any sensitive nature they might have. 
	For instance, given their age and the absence of personal identifiable information, 
	the data present
	very little utility in helping reduce 
	imminent risks to people.
	
	From a regulatory standpoint, in the context of earlier work on online anonymous marketplaces  \cite{christin2013traveling,soska15markets},
	Carnegie Mellon University's 
	Institutional Review Board (IRB) gave us very clear feedback 
	on what is considered human research and thus subject to IRB review. 
	Analyses relying on user-generated content 
	do not constitute human-subject research, 
	and are thus not the purview of the IRB,
	as long as 1) the data analyzed are 
	posted on public fora and were not the result of direct interaction from 
	the researchers with the people posting, 
	2) no private identifiers or personal identifiable information  
	are associated with them, 
	and 3) the research is 
	not correlating different public sources of data to infer private data.\footnote{This position 
		is in line with Title 45 of the Code of Federal Regulations, 
		Part 46 (45 CFR 46), which defines human research.}
	All of these conditions apply to the present study.
	
	

	\section{Conclusion}
	\label{sec:conclusion}
	
	We have worked on the problem of content moderation by detecting and identifying euphemisms. 
	By utilizing the contextual information explicitly, we not only obtain new state-of-the-art detection results, but also discover new euphemisms that are not even on the ground truth list. 
	For euphemism identification, we, for the first time, prove the feasibility of the task and achieve it on a raw text corpus alone, without relying on any additional resources or supervision.

	\section*{Reproducibility}
	
	\ifanonymized
\else
	Our code and pre-trained models are available on GitHub: \url{https://github.com/WanzhengZhu/Euphemism}. 
	\fi
	
	\section*{Acknowledgments}
	\ifanonymized
\else
	We thank our shepherd, Ben Zhao, and the anonymous reviewers for comments 
	on earlier drafts that significantly helped improve this manuscript; 
	Kyle Soska for providing us with the Reddit data and Sadia Afroz for the weapons data;
	Xiaojing Liao and Haoran Lu for availing and discussing the Cantreader 
	implementation with us; and Xin Huang for insightful discussions. 
	This research was partially supported by the National Science Foundation, awards 
	CNS-1720268 and CNS-1814817. 
	
	\fi

	
	
	\bstctlcite{ieeetran:tweaks}
	
	\bibliographystyle{IEEEtran}
	\bibliography{main}
	
	\appendix
	
	\begin{table*}
		\centering
		\caption{Euphemism detection results by our approach (better viewed in color). 
			Purple bold words are correctly detected euphemisms and on the ground truth list (\ie, the DEA list). 
			The purple underlined words indicate that they are incorrect by themselves, but are contained in true euphemism phrases, such as ``dog food", ``Chinese Tobacco" (euphemisms for ``heroin" and ``opium" respectively). 
			Those words which do not appear in the ground truth list are marked black. 
		}
		\includegraphics[width=1\linewidth]{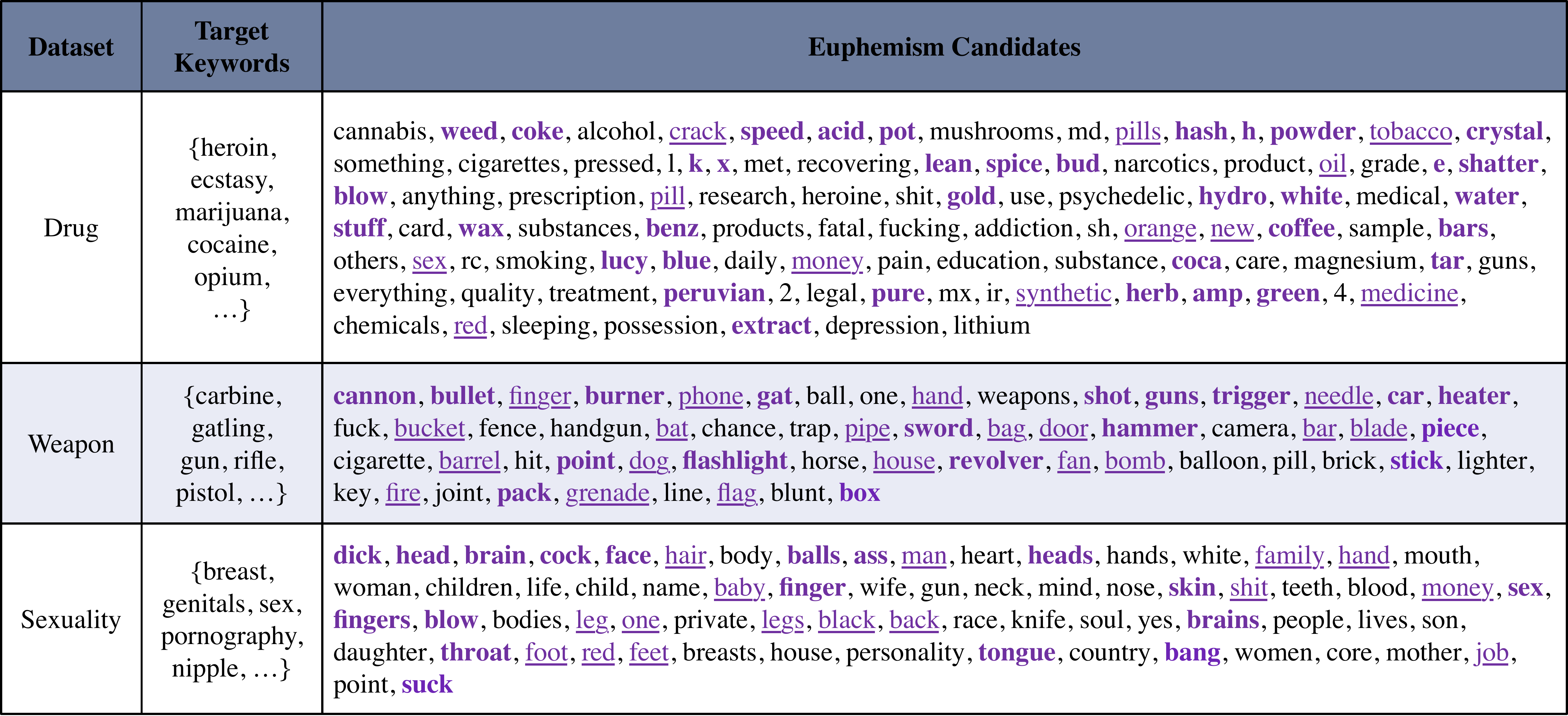}
		\label{fig:casestudies-detection}
	\end{table*}
	
	\begin{table*}
		\centering
		\caption{Case studies of the false positive detection results on the drug dataset. They are real examples from Reddit.}
		\includegraphics[width=0.96\linewidth]{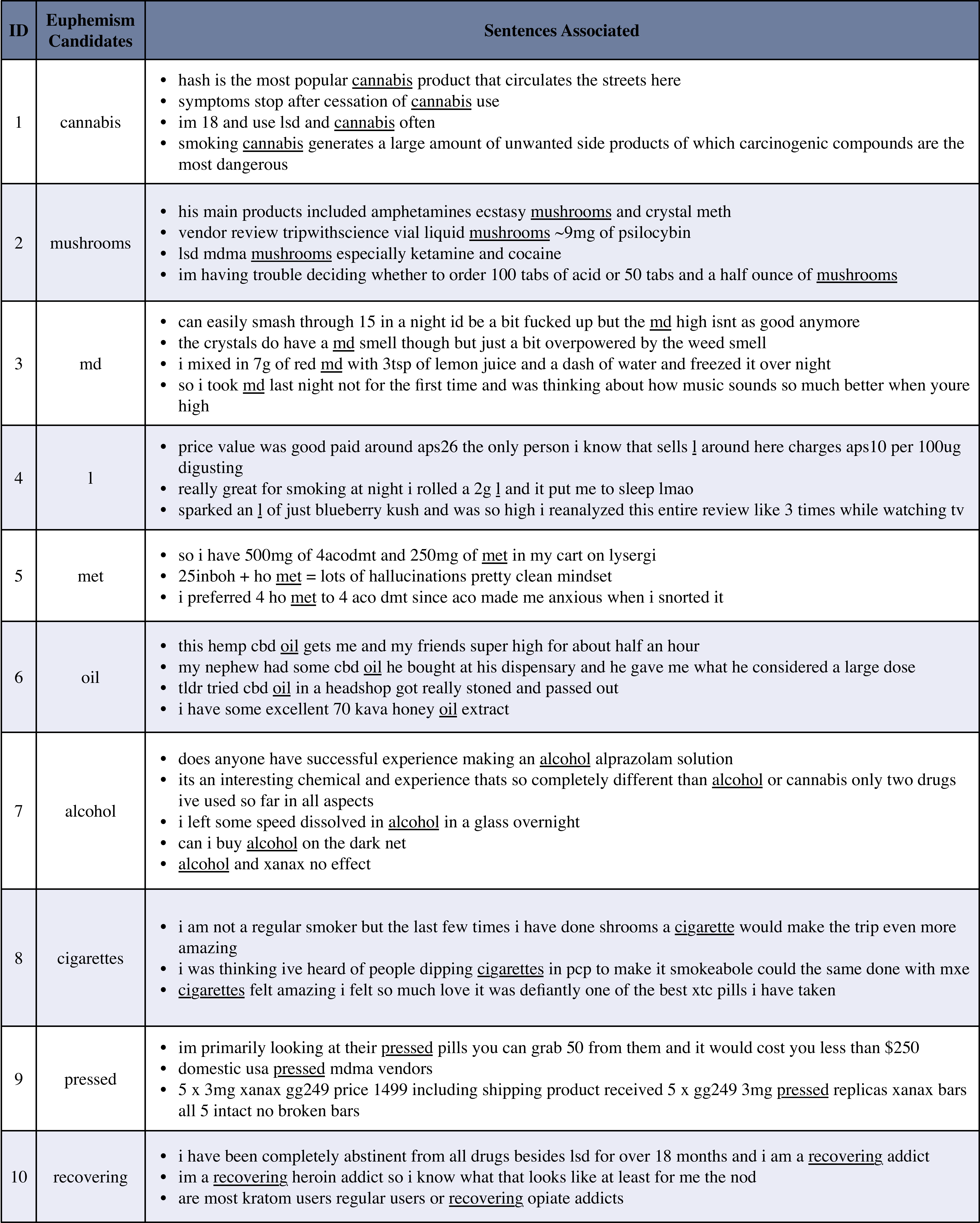}
		\label{fig:casestudies-detection2}
	\end{table*}
	
	\section{Case Study of Euphemism Detection}
	\label{sec:appendix}
	
	We present the euphemism detection results by our approach in Table \ref{fig:casestudies-detection} and analyze the false positive detection results on the drug dataset in Table \ref{fig:casestudies-detection2}. 
	We categorize our false detection results into four types: 
	\begin{itemize}
		\item They are correct euphemisms but missed on the ground truth list (cases 1-5 in Table \ref{fig:casestudies-detection2}). 
		\item They are not euphemisms by themselves, but they are contained in euphemism phrases. For example, as shown in case 6 in Table \ref{fig:casestudies-detection2}, ``oil'' is not a drug euphemism while ``cbd oil'' is one. 
		\item Though they are not euphemisms, they are strongly related to drug or the usage of drug (cases 7-10 in Table \ref{fig:casestudies-detection2}). Cases 7 and 8 uncovers some ways that people take drugs (together with alcohol or cigarettes).
		\item Incorrect detection. 
	\end{itemize}
	
	The case studies reveal that we can even find some correct euphemisms that are not on the ground truth list, which suggests the rapid-evolving nature of euphemisms and the necessity of the automatic euphemism detection task.

\end{document}